\documentclass{article}

\usepackage{amsmath}
\usepackage{microtype}
\usepackage{graphicx}
\usepackage{enumitem}
\usepackage{booktabs} %
\usepackage{amsthm}

\usepackage{hyperref}
\usepackage[capitalise]{cleveref}
\usepackage{rotating}
\usepackage{multirow}

\newcommand{\isep}{\mathrel{{.}\,{.}}\nobreak}

\newcommand{\themethod}{NCoRE}
\newcommand{\thetitle}{\themethod: Neural Counterfactual Representation Learning\\ for Combinations of Treatments}

\theoremstyle{assumption}
\newtheorem{assumption}{Assumption}[]

\usepackage[accepted]{arxiv-icml2018}

\icmltitlerunning{Neural Counterfactual Representation Learning for Combinations of Treatments}
\DeclareUnicodeCharacter{0080}{ }

\begin{document}

\twocolumn[
\icmltitle{\thetitle}

\icmlsetsymbol{equal}{*}

\begin{icmlauthorlist}
\icmlauthor{Sonali Parbhoo}{equal,seas}
\icmlauthor{Stefan Bauer}{mpi}
\icmlauthor{Patrick Schwab}{equal,gsk}
\end{icmlauthorlist}

\icmlaffiliation{seas}{John A. Paulson School of Engineering and Applied Sciences, Harvard University,
Cambridge, USA}
\icmlaffiliation{mpi}{Max Planck Institute for Intelligent Systems, T\"ubingen, Germany}
\icmlaffiliation{gsk}{GlaxoSmithKline, Artificial Intelligence \& Machine Learning, Switzerland}

\icmlcorrespondingauthor{Patrick Schwab}{patrick.x.schwab@gsk.com}

\icmlkeywords{Counterfactual Inference,Combination Treatments, Healthcare}

\vskip 0.3in
]

\printAffiliationsAndNotice{\icmlEqualContribution} %

\begin{abstract}

Estimating an individual's potential response to interventions from observational data is of high practical relevance for many domains, such as healthcare, public policy or economics. In this setting, it is often the case that combinations of interventions may be applied simultaneously, for example, multiple prescriptions in healthcare or different fiscal and monetary measures in economics. However, existing methods for counterfactual inference are limited to settings in which actions are not used simultaneously. Here, we present Neural Counterfactual Relation Estimation (\themethod), a new method for learning counterfactual representations in the combination treatment setting that explicitly models cross-treatment interactions. \themethod{} is based on a novel branched conditional neural representation that includes learnt treatment interaction modulators to infer the potential causal generative process underlying the combination of multiple treatments. Our experiments show that \themethod{} significantly outperforms existing state-of-the-art methods for counterfactual treatment effect estimation that do not account for the effects of combining multiple treatments across several synthetic, semi-synthetic and real-world benchmarks.
\end{abstract}

\section{Introduction}
\begin{figure}[t!]
\begin{centering}
    \vspace{-0.3em}
    \includegraphics[width=0.9\columnwidth]{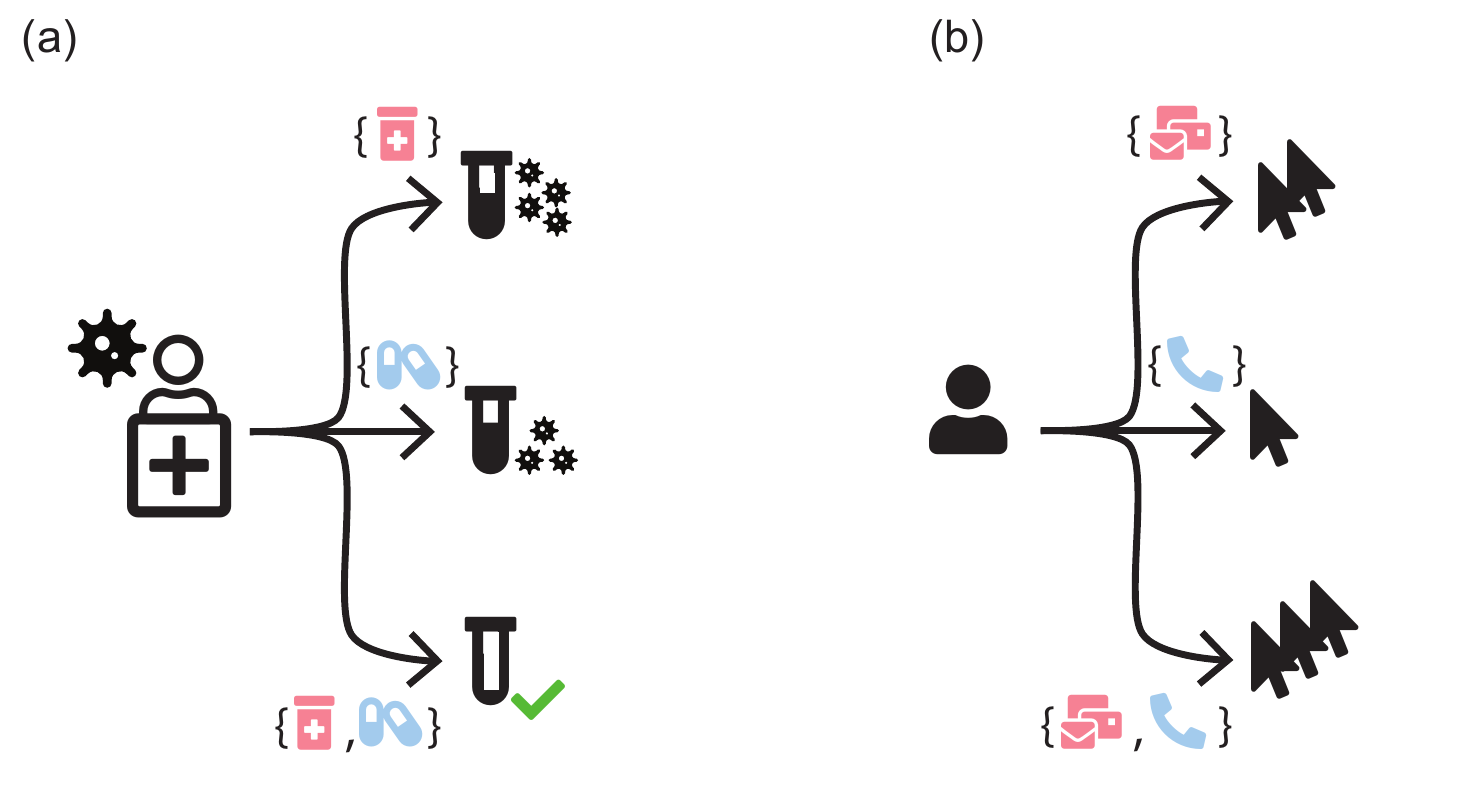}
    \vspace{-1.1em}
    \caption{We study counterfactual teatment effect estimation for combinations of treatments. The presented setting is of high practical importance in various disciplines such as for example healthcare (left, a) and advertisement (right, b) - where one typically has multiple potentially synergistic intervention options available. In the healthcare example, a potential problem setting is the choice between different combination therapies for human immunodeficiency virus (HIV), or, in the marketing example, to present different product placements to a potential customer. Existing methods do not model the relationship between multiple treatments applied concurrently as we propose here.
}
    \label{fig:abstract_vizualization}
\end{centering}
\end{figure}

Understanding the causal effects of an intervention is a key question in many applications, such as personalised medicine (e.g. \citet{wager2017estimation, AlaaS17}), advertisement \citep{bottou2013counterfactual}, or in education \citep{zhao2017estimating} or economics \citep{athey2017state}. While in some applications, only one intervention is possible at a time or actions can be performed sequentially, especially in healthcare applications such as polypharmacology, the concurrent use of \emph{multiple interventions} is necessary. As a concrete example, consider the case of treating patients with human immunodeficiency virus (HIV), where the standard of care is to prescribe synergistic combinations of antiretrovirals concurrently to prevent viral escape. Similarly, again in the healthcare setting, patients may be affected by multiple co-morbidities who may have to be addressed concurently with multiple prescriptions. In economics, multiple fiscal and monetary measures might have to be applied at the same time, while in digital marketing multiple ads are typically shown at the same time on websites. In Figure \ref{fig:abstract_vizualization}, we outline two illustrative application scenarios in healthcare and marketing, respectively. 

Whilst existing methods are focused on estimating the effects of continuous or categorical treatments, we examine the setting where a unit has a set of potentially multiple \emph{concurrent} treatments applied. This setting is particularly challenging for learning algorithms since, for example, for 15 available treatments one would need to consider $2^{15}=32\text{ }768$ possible combinations of which many may not have been applied in the available observational data. This combinatorical explosion of potential treatment combinations necessitates an expressive and effective modelling approach that scales to the large number of potentially permissible cross-treatment interactions.

To address this challenging setting, we introduce Neural Counterfactual Relation Estimation (\themethod), a novel method for learning meaningful counterfactual representations for combinations of treatments. Our  approach is based on a new conditional neural representation containing treatment interaction modulators that model the way in which combinations of treatments may interact at a time. On several synthetic, semi-synthetic and real-world benchmarks, \themethod{} significantly outperforms existing methods for counterfactual treatment effect estimation that do not account for the effects of combining multiple treatments. 

\textbf{Contributions.} We present the following contributions:
\begin{itemize}[noitemsep]
\item We introduce \themethod{}, a novel approach for training neural networks for counterfactual inference that, in contrast to existing methods, is suitable for estimating the effects for combinations of treatments. 
\item We extend several existing approaches for treatment effect estimation from the binary to the multiple treatment setting.  In contrast to existing approaches and na\"ive extensions thereof,  \themethod{} leverages relationships between interventions and is able to extrapolate to unseen combinations in real-world situations (presented in an evaluation on experimental CRISPR-Cas9 Three-way Knockout data). 
\item We perform extensive experiments on synthetic, semi-synthetic and real-world data settings. The experiments show that \themethod{} significantly outperforms a number of more complex state-of-the-art methods for counterfactual treatment effect estimation that do not account for the effects of combining multiple treatments. 
\end{itemize}
\section{Related Work}
\label{sec:related_work}
\textbf{Background.} Inferring the causal effects of interventions is a central pursuit in many important
domains, such as healthcare, economics, and public policy. In medicine, for example, treatment
effects are typically estimated via rigorous prospective studies, such as randomised controlled trials (RCTs), and their results are used to regulate the approval of treatments. However, in many settings of interest, randomised experiments are too expensive or time-consuming to execute, or not possible for ethical reasons \citep{carpenter2014reputation, bothwell2016assessing}. Observational data, i.e. data that has not been collected in a randomised
experiment, on the other hand, is often readily available in large quantities. In these situations,
methods for estimating causal effects from observational data are of paramount importance. This is an important but challenging problem since the data is likewise often confounded. 

\textbf{Propensity Score-based Methods.} Since their introduction, propensity scores \cite{propintro} have been widely used to control for pre-treatment imbalances on observed variables in observational studies. Several approaches use these scores to perform matching\cite{stuart2010matching,rosenbaum2002overt,kallus2020generalized}, stratification \cite{rosenbaum1984reducing} or weighting of samples from similar treatment and control groups \cite{mccaffrey2004propensity} to reduce the effects of confounding. They offer several advantages to standard regression-based covariate adjustment techniques and have thus also been extended to settings with multiple treatments of interest. Specifically, \citet{imbens2000role, robins2000marginal, imai2004causal, mccaffrey2013tutorial} and \citet{lopez2017estimation} use inverse probability of treatment weighting (IPTW) for modelling treatment effects in multi-treatment settings. Further works by \citet{lechner2001identification}, \citet{zanutto2005using} and \citet{spreeuwenberg2010multiple} demonstrate application of propensity scores for matching and stratification. However, none of these works provides practical guidance on the use of propensity scores in a multi-treatment setting, nor are any of these methods applicable for use with multiple concurrent treatments as we propose here.

\textbf{Latent Variable Models for Estimating Treatment Effects.} 
Deep latent variable models such as \citep{Louizos,JohanssonTAR,yoon2018ganite,bica2020time} are popular tools for individualised treatment effect estimation. These methods all learn latent representations on the basis of observed confounders or noisy proxies, and condition on these representations for inferring treatment outcomes. Alternative methods such as \citet{parbhoo2017combining} model the effects of a series of treatments as a Markov Decision Process, where treatments are treated as independent actions and can influence one another only via changes to the state. Additional work from \citet{schwab2020learning} also explores how to reason about effects in continuous settings, where different dosages of treatments produce different patient outcomes. In contrast to these methods, \themethod{} focuses on estimating effects of multiple \emph{concurrent} treatments where we explicitly model the cross-interactions between treatments. 

\textbf{Treatment Combinations.} 
Several other works examining the effects of combinations of treatments \citep[e.g.][]{pmlr-v126-parbhoo20a} have been proposed from a reinforcement learning perspective where each treatment combination constitutes a different action. Unlike these methods, our work explicitly focuses on modelling the interactions between various treatments to deduce their combined effects. Similar causal interaction methods for application in conjoint analysis have been explored in \cite{luce,green}. Recently, \citet{egami2018causal} also studied the problem of causal interaction in factorial experiments, where several factors at differing levels may be randomised to produce a large number of treatment combinations. Specifically, the authors focus on estimating a metric known as the Average Marginal Interaction Effect (AMIE) to reason about the choice of treatments. The authors uses a penalised regression model this purpose. Unlike these studies, the work we present in this paper focuses on estimating individual treatment effects rather than average treatment effects and introduces a new conditional neural architecture containing treatment interaction modulators to model cross-treatment interactions.

\begin{figure}[t!]
\centering
    \includegraphics[width=0.95\columnwidth]{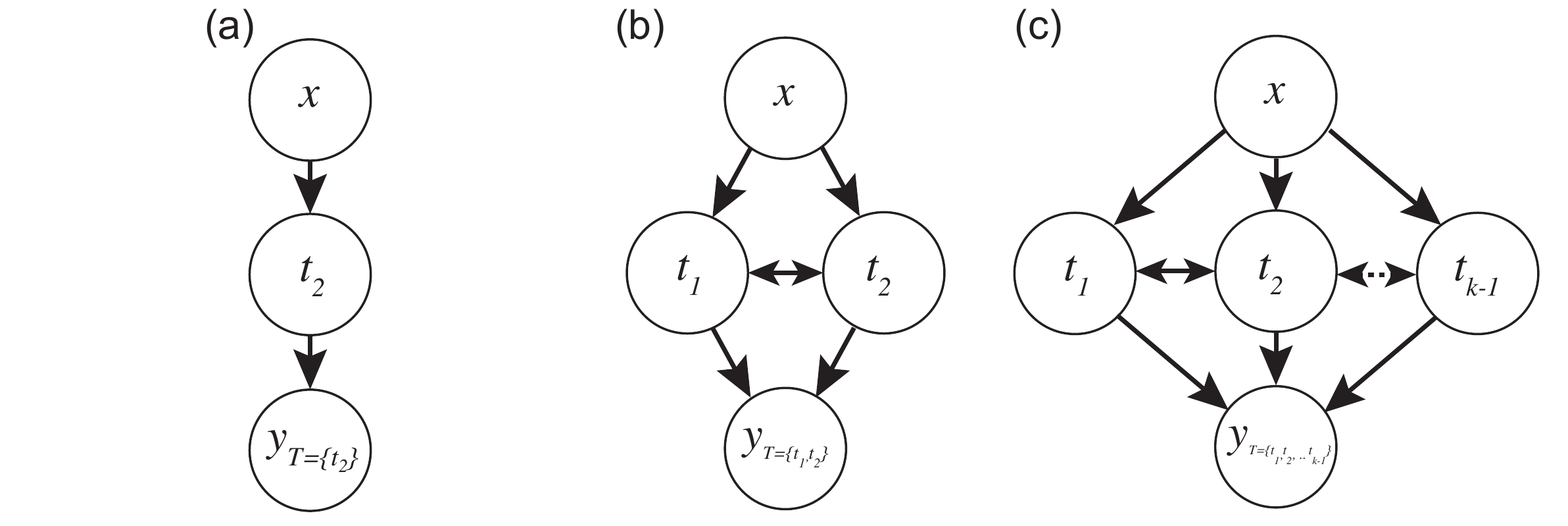}
    \vspace{-0.5em}
    \caption{Three examplary directed acyclic graphs (DAGs) in which (a) only one treatment is applied (left), (b) two treatments are applied concurrently (centre) or (c) all treatments are applied concurrently (right). Note the potential for cross-treatments interactions when multiple treatments are applied concurrently.}
\label{DAG}
\end{figure}

\section{Methodology} 
\label{sec:method}
\textbf{Problem Setting.} We consider a setting where we are given $n$ observed samples, or units, $x$ with pre-treatment covariates $x_i$ for $i \in [1 \isep p]$ with $p$ being the number of available per-unit covariates. For each sample, the potential outcomes $\hat{y}_{T}$ are the responses to a subset of treatments $T$ where constituent treatments $t_j$ are chosen out of the complete available set of treatments $T_\text{complete} = \{t_0, t_1, \ldots, t_{k-1}\}$ with $j\in [0 \isep k-1]$ and $k$ being the number of available treatment options. For training, we receive observed samples $x$ with outcomes $y_{T}$ for only one particular set of treatments $T$ that may be different for each sample. We subsequently use the training data to train a model to produce counterfactual estimates $\hat{y}_{{T}}$ of the potential outcome after applying treatment set $T$. This corresponds to the Rubin-Neyman potential outcomes framework extended to multiple treatments that may be applied concurrently. Three causal directed acyclic graphs (DAGs) exemplifying this setting are presented in Figure \ref{DAG}. Note that, depending on the modelling context, not all possible combinations of treatments in $T_\text{complete}$ may be permissible. For example, the empty treatment set $T=\{\}$ considering the effect of non-intervention may not in all settings be an available option.

\begin{figure*}[t!]
\centering
    \vspace{-0.65em}
    $\vcenter{\hbox{\includegraphics[width=0.489\linewidth,page=1]{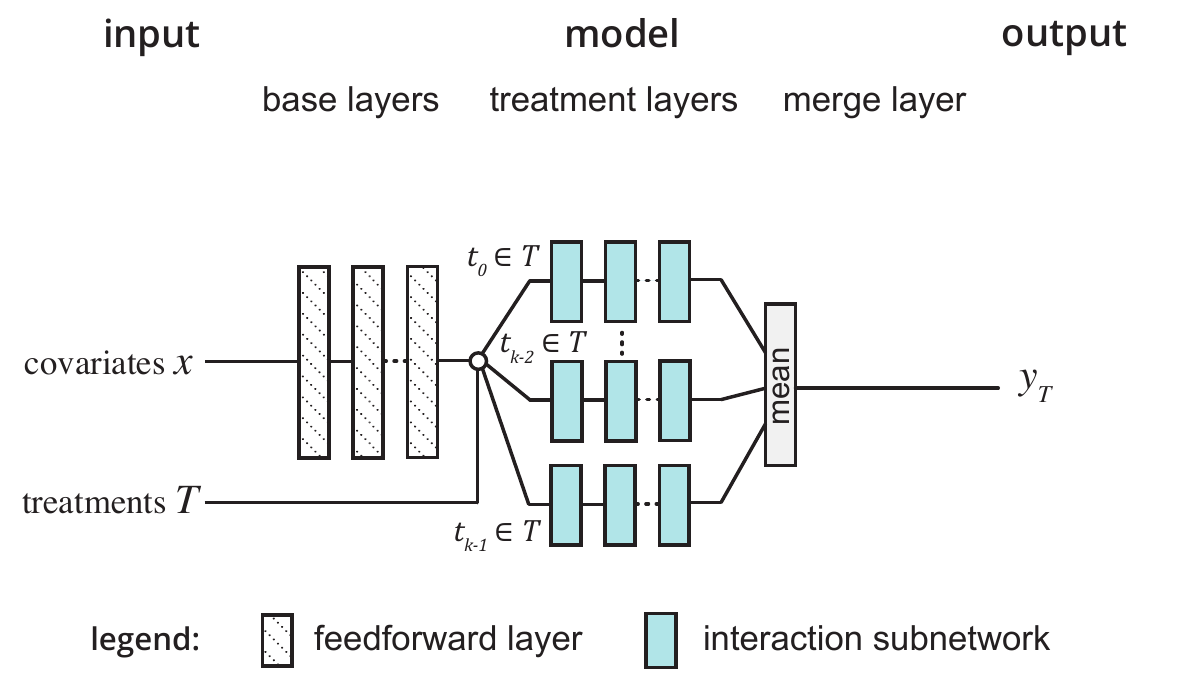}}}$
    $\vcenter{\hbox{\includegraphics[width=0.489\linewidth,page=1]{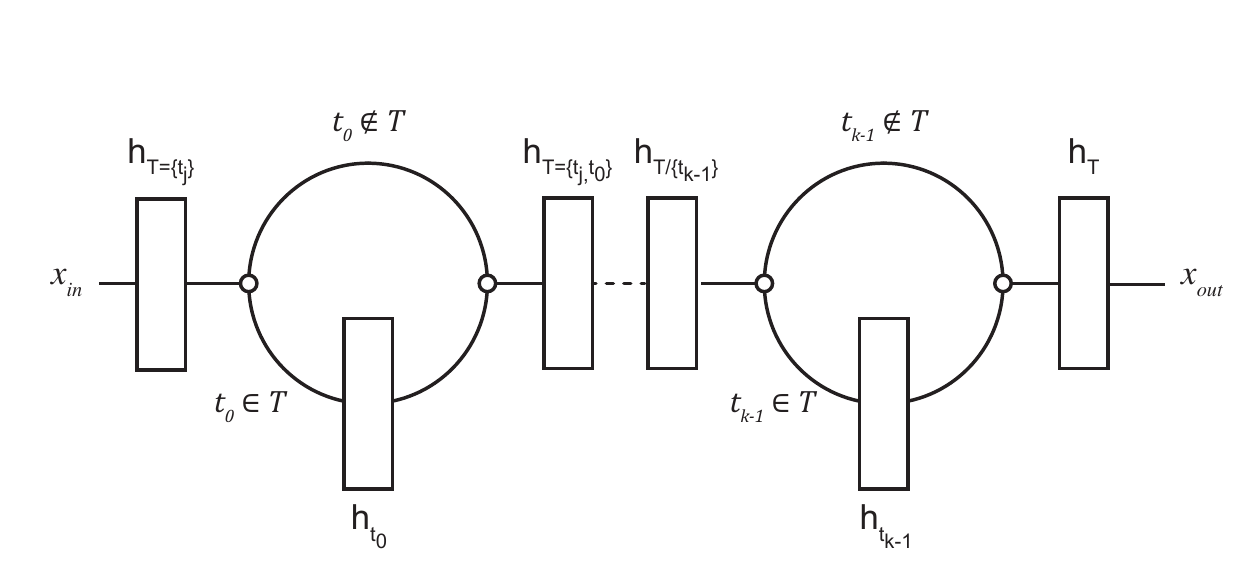}}}$
    \vspace{-0.7em}
    \caption{An overview of \themethod{} (left) and the structure of each of the interaction subnetworks (right) used in \themethod{}'s treatment arms. Structurally, \themethod{} consists of a variable number of shared base layers (striped) with $k$ intermediary treatment layers which are then merged to obtain a predicted outcome $y_T$.  The shared base layers are trained on all samples and serve to model cross-treatment interactions, and the treatment layers are only trained on samples from their respective treatment category and serve to model per-treatment interactions. The key part how combinations of treatments are modeled in \themethod{} is through the interaction subnetworks (left in light blue) contained within each treatment layer. The $j$th treatment layer is only trained on the units which received the respective treatment option $t_j$ in their set of observed treatments $T$. The interaction subnetworks (right) used within \themethod{}'s treatment arms have a sequential, conditionally controlled structure that enables them to model the interaction of concurrent treatments - with each interaction having multiplicative and/or additive influence on the base learnt hidden representation $h_{T=\{t_j\}}$ in sequence until the final output hidden representation $h_T$ depending on whether the $j$th treatment was included in the respective unit's set of treatments $T$.}
\label{fig:overview}
\end{figure*}

We make the following standard assumptions in order to infer treatment effects from observational data:
\begin{assumption}{\bf{(Ignorability.)}} We assume unconfoundedness i.e. the distribution of the treatment is independent of potential outcomes when given the observed variables. 
\end{assumption}
\begin{assumption} {\bf{(Overlap).}} Every unit has non-zero probability of receiving either treatment given their observed covariates.
\end{assumption}
\begin{assumption}{\bf{(Stable Unit Treatment Value.)}}
We assume the distribution for the potential outcomes of a particular unit are unaffected by treatment assignments of another unit when given the covariates $X_i$. 
\end{assumption}

\textbf{Neural Counterfactual Relation Estimation (\themethod).} 
An important design consideration when performing counterfactual inference is model structure \cite{JohanssonTAR,AlaaS17}. In the combination treatment setting, we would ideally aim to explicitly model interactions between various treatments such that the effects of applying combinations of concurrent interventions can estimated. We therefore propose the use of a Neural Counterfactual Relation Estimation (\themethod) model, a fully differentiable neural network model that consists of a variable number of base layers that operate on the unit covariates $x$ followed by $k$ treatment arms, one for each base treatment $t_j$ with a variable number of per-treatment interaction subnetwork layers (\Cref{fig:overview}). During training, shared base layer parameters are updated for every observed sample, whereas the parameters of a treatment arms are only updated if the respective treatment $t_j$ is part of the treatment set $T$ applied to that sample. Mathematically, the interaction subnetworks $h_T$ operating on an input $x_in$ (either passed from the shared base layers or another subnetwork depending on location) that make up the base layers of the treatment arms are recursively defined as:
\begin{align*}
h_T(x_{in}) = \begin{cases}
      W_0x_{in}+b_0, & \text{if}\ T=\{\} \\
      W_{t_j}h_{T\setminus\{t_j\}}(x_{in})+b_{t_j}, & \text{if}\ t_j \in T
    \end{cases}
\end{align*}
With $W_0$ and $W_{t_j}$ and $b_0$ and $b_{t_j}$ being weight matrices, biases of the base feedforward layers and the per-treatment interaction submodules that are jointly optimised with the other model parameters during training for each available treatment $t_j$ respectively. Using this formulation, \themethod{} ensures concurrent treatment arms all sequentially influence the learnt hidden representation and thereby explicitly models potential cross-treatment interactions (see \Cref{fig:overview} for an unrolled illustration). In contrast to \themethod{}, existing methods that, for example, na\"ively utilise separate models for each treatment combination tend to suffer from data sparsity since only units that receive a specific treatment combination of may be used to train each arm. As the number of treatment combinations grows, there may not be adequate samples available to train a separate combination model. \themethod{} overcomes this issue by enabling information to be shared across treatments through the base layers of our architecture and by explicitly modelling cross-treatment interactions via the interaction subnetworks.

\textbf{Treatment Assignment Bias.} Beyond structural deliberations, another important factor in learning counterfactual models is treatment assignment bias. Treatment assignment bias refers to systematic differences in the covariate distributions of observed unit samples between groups treated with different sets of treatments $T$. Such observed or hidden confounding can preclude the disentanglement of causal effects between differences in outcomes that can be attributed to differences in the underlying covariate distribution and differences in outcomes that can be attributed to the actual effect of treatment set $T$. As outlined in Section \ref{sec:related_work}, in the well-studied setting with only two available treatments that may not be used concurrently, researchers have proposed the use of balancing scores  \cite{rosenbaum1984reducing,stuart2010matching,rosenbaum2002overt,kallus2020generalized,schwab2018pm} or distribution matching \cite{JohanssonTAR,pmlr-v70-shalit17a} to address observed confounding under assumptions of no hidden or unobserved confounding. In the combination treatment setting, the direct application of balancing scores is however, in general, computationally not tractable as the number of matched controls that would have to be included grows exponentially with the number of treatment options available in $T_\text{complete}$. We therefore introduce a new randomised algorithm for approximate batch matching to address observed confounding in the combination treatment setting. The algorithm, starting from a randomly selected seed unit's covariates out of the available training data, iteratively builds a batch $B$ consisting of a pre-specified number $s$ of samples that is approximately balanced across the included treatment combinations. At each iteration, the algorithm randomly selects from the remaining training data the closest matching unit with an observed treatment combination also randomly chosen from those not yet represented in the batch $B$. Closest matches are determined by matching on an appropriate balancing score, such as for example a lower dimensional representation of the covariates $x$. We note that, in practice, an appropriate spatial search structure may be used to computationally accelerate the search of closest matches in the balancing score space, and a dimensionality reduction algorithm such as principal components analysis (PCA) may be used to construct a lower dimensional covariate space for matching\footnote{In principle, matching on the propensity score vector \cite{lechner2001identification} instead of a balancing score determined by dimensionality reduction is also an option in the combination treatment setting. However, since the dimensionality of the propensity score vector grows exponentially with the number of treatments in $T_\text{complete}$, using a balancing score based on dimensionality reduction is typically preferrable as the dimensionality of the balancing score is then independent of the number of treatments available.}.

\textbf{Metrics.} In order to enable a meaningful comparison of models in the presented setting, we use metrics that cover desirable aspects of models trained for estimating treatment responses. Specifically, we evaluate models by calculating the root mean squared error (RMSE) 
\begin{align*}
\text{RMSE}(y, \hat{y}) = \frac{1}{2^{k}-1}\Sigma^{j\in [1 \isep 2^{k}-1]} (y_j - \hat{y}_j)^2
\end{align*}
between the true counterfactual outcomes $y$ and the estimated counterfactual outcomes $\hat{y}$ for all $2^k-1$ possible combinations of treatments in the set of all available treatments $T_\text{complete}$ across all patients in the test fold.

\section{Experiments} 
\label{sec:experiments}
To evaluate \themethod{}, we conducted an experimental evaluation that aimed to answer the following questions:
\begin{itemize}[noitemsep]
\item[(1)] How does \themethod{} compare to existing state-of-the-art approaches to inferring counterfactual outcomes in settings with multiple available treatments that can be used in conjunction?
\item[(2)] How is the counterfactual estimation performance of \themethod{} and existing state-of-the-art baselines impacted by varying (i) the number of available treatments, (ii) the number of samples available for training and evaluation, and (iii) the treatment assignment bias with which treatments are assigned to units?
\end{itemize}

\textbf{Datasets.} To empirically answer the above questions, we utilised three datasets consisting of synthetic and semi-synthetic real-world data collected in different environments (\Cref{tb:datasets}). Using the three datasets, we created an open suite of benchmark tasks that are aimed to represent a variety of relevant use cases for counterfactual inference in settings with multiple available treatments that may be combined. %
For all experiments, datasets were randomly split into training (\~{}60\%\footnote{Fold size ratios are approximate because \texttt{scikit-multilearn} may sometimes vary fold sizes slightly to improve the overall fold balance.}), validation (\~{}20\%) and test folds (\~{}20\% of all samples) in strata of key baseline features using the \texttt{scikit-multilearn} package \cite{2017scikitmultilearn} (version 0.2.0).

\textbf{Human Immunodeficiency Virus (HIV) Simulator.} To enable testing of the sensitivity of \themethod{} and other state-of-the-art methods for counterfactual inference to the number of available treatments, the number of samples available, and treatment assignment bias, we constructed a fully simulated setting for modelling treatment response to pharmaceutical interventions targeting patient populations with human immunodeficiency virus (HIV) infections and acquired immunodeficiency syndrome (AIDS). HIV/AIDS treatment is well-suited for evaluating counterfactual estimators that include combination effects because treatment with combination antiretroviral therapy (cART) - a combined cocktail of multiple antiretroviral drugs with different mechanisms of action - is the standard of care for treating HIV \cite{freedberg2001cost}. The design of our simulator consisted of a first stage in which we generated baseline pre-treatment HIV/AIDS patient characteristics $x$, including age, sex, ethnicity, membership in risk groups, AIDS status, prior treatment history, mutation status, and pre-treatment CD4 cell counts and viral load (VL) measures. The first stage of the simulator was designed to generate baseline characteristics of HIV/AIDS patients that match baseline characteristic distributions observed in real-world observational studies. We then, at random and with uniform probability, selected $k$ patient archetypes $a_k$ for each available treatment option that were used to assign $|T_i| \sim \text{Poisson}(2)+1$ with $t \in [1 \isep k]$ treatments to patient $i$ (truncated to a maximum of $k$). To assign treatments to patients, we calculated the distance 
\begin{align*}
\text{MixedDistance}(x_1, x_2) = \\
 \frac{|I_\text{discrete}|}{p}\text{Jaccard}(x_{1,I_\text{discrete}}, x_{2,I_\text{discrete}}) &+\\ \frac{|I_\text{continuous}|}{p}\Sigma^{l \in I_\text{continuous}}|x_{1,l} - x_{2,l}|
\end{align*}
 consisting of the weighted sum of Jaccard and mean absolute distances of the continuous baseline demographics with $I_\text{discrete} = \{q | x_q = \text{discrete}\}$ and $I_\text{continuous} = \{q | x_q = \text{continuous}\}$ respectively as the sets of indices of discrete and continuous covariates in $x$ between each patient's pre-treatment covariates $x$ and each treatment archetype $c_k$. Treatment assignments for treatment $j$ and a patient with covariates $x$ were then defined as $t_j \sim \text{Uniform}(0, 1) > p_{t_j}$ with $p_{t_j} = \text{softmax}(\kappa{}d(x, a_j))$  as the assignment treatment probability for treatment indicator $t_j \in {0,1}$ for treatment $j \in [0 \isep k-1]$ and with $\kappa$ as the variable treatment assignment bias strength coefficient. $\kappa = 0$ corresponds to no treatment assignment bias and higher numbers correspond to stronger assignment biases based on the patient covariates $x$. We generated counterfactual outcomes $y_{T}$ for each possible combination $T$ of treatments using a two-step process with the first step consisting of sampling a Gaussian outcome model $M_j$ that defines the outcome $y_{T=\{t_j\}}$ for each treatment $j$ used by itself. The per-treatment Gaussian outcome models $M_j$ calculated outcomes for a patient with pre-treatment covariates $x$ using
\begin{align*}
y_{T=\{t_j\}} = \hat{y}_{T=\{t_j\}}\cdot\text{MixedDistance}(x, c_l)
\end{align*}
with unscaled outcomes $\hat{y}_{T=\{t_j\}} \sim \mathcal{N}_\text{truncated}(\mu_{c_l}, \sigma_{c_l})$ truncated to $\hat{y}_{T=\{t_j\}} \in (L_\text{initial,min}, L_\text{initial,max})$, mean $\mu_{c_l} \sim \text{Uniform}(L_\text{initial,min}, L_\text{initial,max})$, standard deviation $\sigma_{c_l} \sim 0.5$, minimum and maximum initial viral loads $L_\text{initial,min} = 0.84 \frac{1}{\text{ml}}$ and $L_\text{initial,max} = 7.69 \frac{1}{\text{ml}}$ respectively and centroids $c_l = x_l$ with centroid indices $l$ selected at random from the entire training set $l \sim \text{Uniform}(0, N-1)$ consisting of $N$ patients. The second step in the outcome generating process then generated outcomes 
\begin{align*}
y_{T} = \mathcal{P}_{d}(T)\mathcal{P}_{d}(y_{T=\{t_j\}})\cdot{}B^T
\end{align*}
for all possible combinations $T = \{j | t_j = 1\}$ of assigned base treatments $t_j$ with the ordered set of all possible polynomial interactions $\mathcal{P}_{d}(\theta)$ of degree $d = \text{min}(|T|, 5)$ with coefficients $\theta$, combination coefficients $B_o \sim s_{B_o}\cdot\mathcal{N}(1.02^{d_o-1}\mu_B, 1.02^{d_o-2}\sigma_B)$, scarcity coefficient $s_{B_o} \sim (\text{Uniform}(0,1) < 0.2)$, means of combination coefficients $\mu_B = -0.03$, standard deviations of combination coefficients $\sigma_B = 0.015$ and degree $d(\cdot)$ of the interaction $\mathcal{P}_{d}(T)_o$ at index $o \in [0 \isep |\mathcal{P}_{d}(T)|-1]$. %

\begin{table}[t!]
\centering
\vspace{-1.25ex}
\caption{Comparison of the datasets used in our experimental evaluation. We evaluated one synthetic and two semi-synthetic datasets with varying numbers of samples, features and treatments.}
\label{tb:datasets}
\vspace{1.0ex}
\begin{small}
\begin{tabular}{l@{\hskip 2.0ex}r@{\hskip 0.8ex}r@{\hskip 0.8ex}r@{\hskip 2.8ex}r@{\hskip 1.5ex}r}
\toprule
Dataset & Samples & Features & Treatments & Type \\
\midrule
Simulator & any & 2870 & any & synthetic\\
Europe 1 & 3161 & 266 & 20 & semi-synthetic \\
Europe 2 & 3116 & 105 & 18 & semi-synthetic\\
CRISPR KO & 3185 & 2424 & 15 & experimental\\
\bottomrule
\end{tabular}
\end{small}
\end{table}

\textbf{EuResist Integrated Database Cohorts (Europe 1 and Europe 2).} Europe 1 is a subset of data from HIV-infected individuals between 18 and 72 years of age obtained from the EuResist Integrated Database \citep{EuResist}. The database contains the genotype (mutation) data, CD4 count, viral load and therapy data for individuals that have been administered HIV therapies since 1988. The therapy data records the number of prior treatment lines a patient was administered and treatment history. We focus on a subset of these patients that have at least 3 prior treatment lines. Europe 2 is a set of data from Antiretroviral Resistance Cohort Analysis (ARCA) \cite{ARCA} database with the same set of features. For both Europe 1 and Europe 2, treatment assignments and outcomes (viral load after treatment) were generated analogously to the Simulator dataset using a semi-synthetic approach based on the real-world observed pre-treatment covariates $x$ and the observed treatments as treatment assignments.  

\textbf{CRISPR Three-way Knockout (CRISPR KO).} The CRISPR Thee-way Knockout (CRISPR KO) benchmark dataset consists of real-world experimental data collected in a systematic multi-gene knockout screen conducted by \citet{zhou2020threeway}. In their screen, \citet{zhou2020threeway} used a CRISPR-Cas9-based multi-gene knockout system to experimentally quantify the synergistic or antagonistic causal genetic interaction of a total of 3185 combinations (up to three-way interactions) of 15 genes $T=\{\text{CDK4}$, $\text{DNMT1}$, $\text{EGFR}$, $\text{ERBB2}$, $\text{FGF2}$, $\text{HDAC1}$, $\text{HPRT1}$, $\text{MAP2K1}$, $\text{MTOR}$, $\text{PIK3C3}$, $\text{POLA1}$, $\text{TGFB1}$, $\text{TOP1}$, $\text{TUBA1A}$, $\text{TYMS}$, $\text{VKORC1}\}$ that are targets of existing ovarian cancer treatments in a cell line representative for high-grade serous ovarian cancer (HGSOC). To measure the causal effect $y_T$ of targeting a set $T$ of genes, \citet{zhou2020threeway} compared the log$_2$ fold change in barcoded guide ribonucleic acid (gRNA) counts in cells cultured for 15 and 26 days as a measure for the degree to which cancer cell growth was modulated by the knockout - where an increased inhibition of cell growth was assumed to be associated with a potential synergistic causal effect of targeting the respective genes using a combination therapy. To estimate the relationship between sets of genes and the causal effect of targeting those genes on inhibiting ovarian cancer cell growth in vitro, we used a feature set $X = [X_{G_1}, X_{G_2}, X_{G_3}]$ consisting of three 808-dimensional gene descriptors $X_{G_1}, X_{G_2}, X_{G_3}$ where each gene descriptor contained the as determined using data from genome-wide single knockout CRISPR-Cas9 essentiality screens in 808 cell lines \cite{dempster2019extracting} and computed using the CERES algorithm \cite{meyers2017computational} for genes $G_1$, $G_2$, and $G_3$ included in the set $T$ of genes that were intervened on\footnote{The data is freely available from the DepMap project at \url{https://ndownloader.figshare.com/files/25494359} \cite{depmap202020q4}.}. $X_{G_3}$ and $X_{G_2}$ and $X_{G_3}$ were zero-imputed to indicate missingness for single or two-way gene knockouts with a total of only one or two genes involved, respectively. The CRISPR KO benchmark is particularly challenging for causal inference methods because of the complex underlying biological process that determines causal effects on cancer cell growth in a realistic model system of a heterogenous disease, the large number of potential treatment combinations, high-dimensional covariate space, and the sparsity of labelled data available (at most one outcome for each combination). Due to only one sample being available for each evaluated gene combination, the CRISPR KO benchmark also tests the ability of counterfactual inference methods to extrapolate beyond treatment sets $T$ observed in the training data, as the randomly split off test set consisted entirely of novel combinations not present in the training or validation data, i.e. it could, for example, tests the ability of counterfactual estimators to estimate the causal effect of intervening on genes $\{G_1, G_2, G_3\}$ based on data observed after intervening on genes $\{G_1\}$, $\{G_3\}$, $\{G_2, G_3\}$, and experimental data from other genes with potentially similar gene descriptors $X$.

\begin{table}[b!] 
\caption{Hyperparameters. \textup{Hyperparameter ranges used for hyperparameter optimisation of Ridge Regression (RR), Deconfounder (DC) and Neural Network (NN; includes GANITE, TARNET, and \themethod{}) models for all benchmark tasks. Comma-delimited lists indicate discrete choices sampled with equal selection probability.}}
\vspace{1.0ex}
	\label{tb:hyperparameters}
	\centering
	\begin{small}
		\begin{tabular}{l@{\hskip 1.25ex}l@{\hskip 10.5ex}r}
			\toprule
			& Hyperparameter & Choices\\
			\midrule
			\parbox[t]{2mm}{\multirow{1}{*}{\rotatebox[origin=c]{90}{RR}}}&Regularization strength $C$ & 0.1, 1.0, 10.0\\
			\midrule
			\parbox[t]{2mm}{\multirow{1}{*}{\rotatebox[origin=c]{90}{DC}}}&Latent dimension $D$ & 3, 5, 10, 15\\
			\midrule
			\parbox[t]{2mm}{\multirow{7}{*}{\rotatebox[origin=c]{90}{NN}}}& Number of hidden units $N$ & 8, 16, 32, 64\\
			& Number of layers $L$ & 1, 2, 3\\
			& Batch size $B$ & 16, 32, 64, 128\\
			& L$2$ regularisation $\lambda_2$ & 0.0, 0.00001, 0.0001\\
			& Learning rate $\alpha$ & 0.003, 0.03\\
			& Dropout percentage $p$ & 0, 10, 15, 25\%\\
			\bottomrule
		\end{tabular}
	\end{small}
\end{table}

\begin{figure*}[ht]
    \vspace{-0.45em}
    \includegraphics[width=0.226\linewidth,page=1]{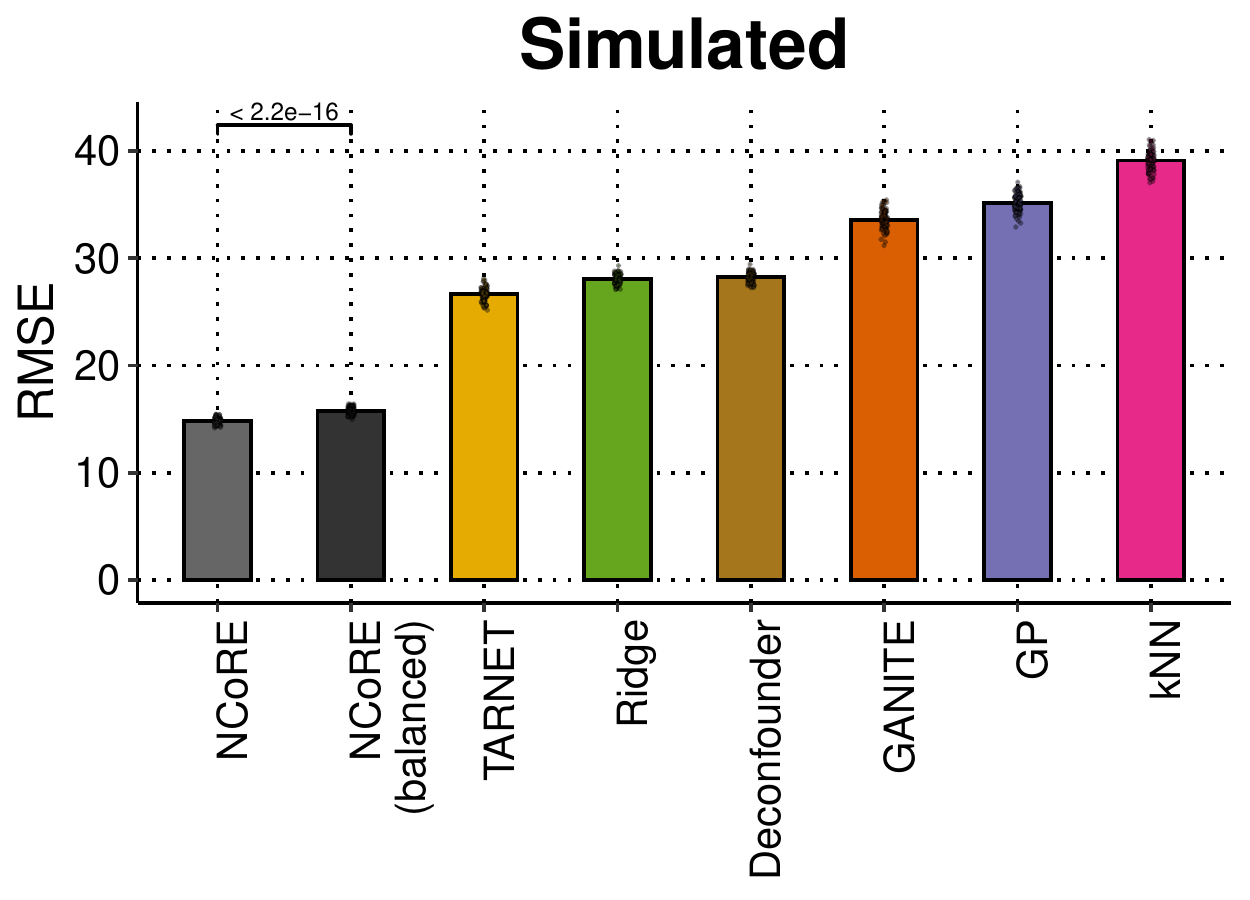}\quad
    \includegraphics[width=0.226\linewidth,page=1]{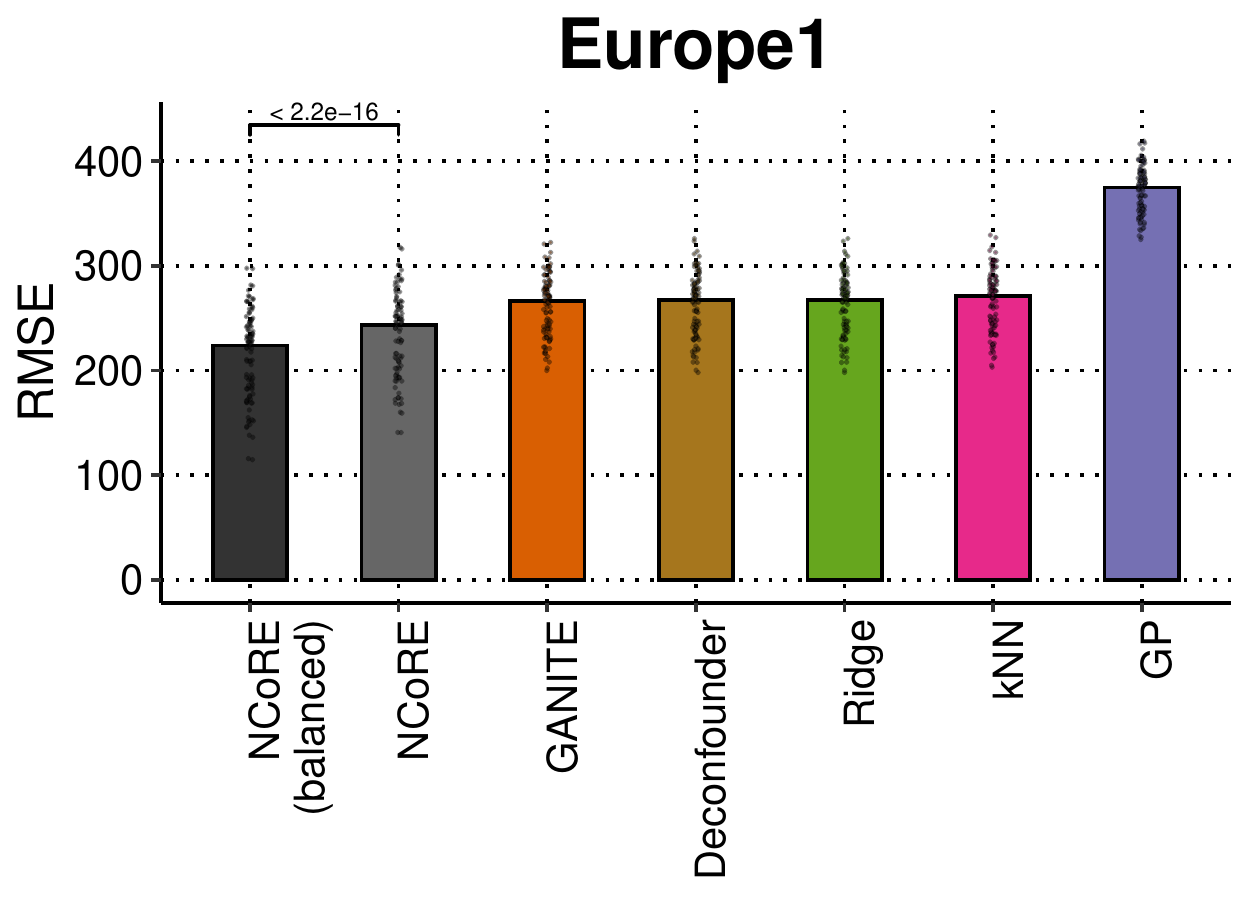}\quad
    \includegraphics[width=0.226\linewidth,page=1]{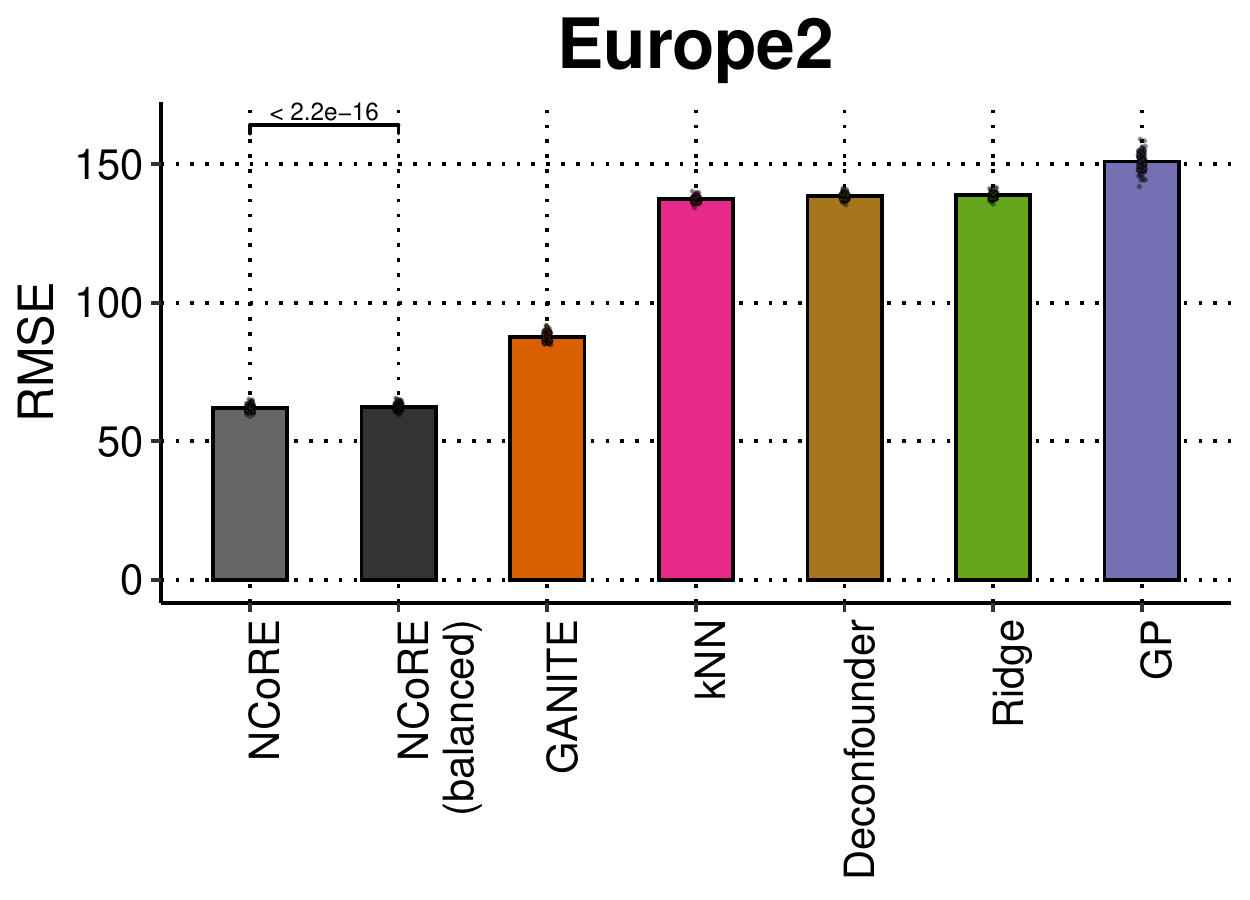}\quad
    \includegraphics[width=0.226\linewidth,page=1]{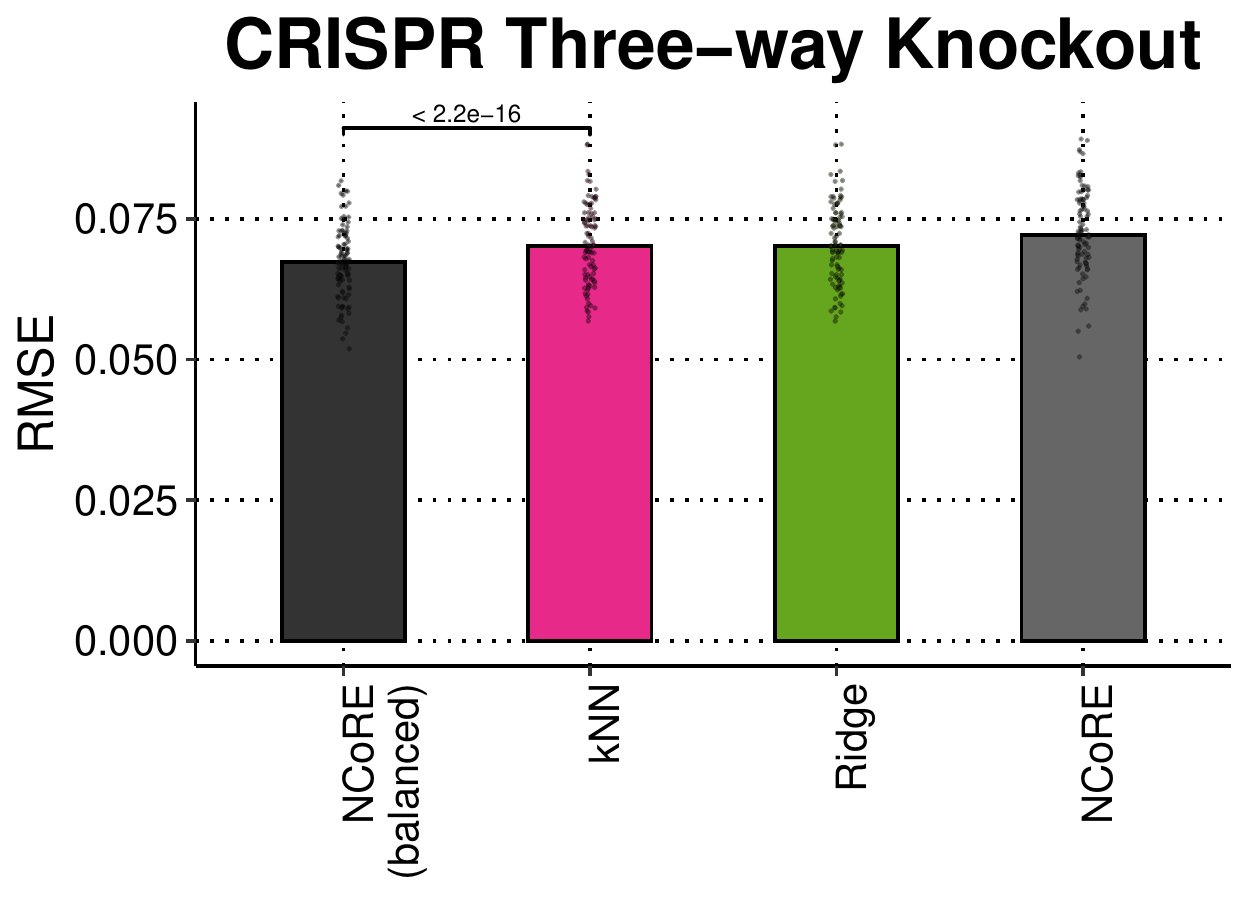}
    \vspace{-1.6em}
    \caption{Performance comparison in terms of RMSE (y-axis) of several state-of-the-art methods (x-axis) for counterfactual treatment across various synthetic (left), semi-synthetic (middle, Europe1 and Europe2) and real-world datasets (right). For simulated data and semi-synthetic data \themethod{} outperforms all competitive baselines in most cases with a significant margin. For the real-world experiment (CRISPR Three-way Knockout), less baselines are shown since due to the computational complexity most baseline approaches did not converge within reasonable time.   }
\label{fig:performances_1}
\end{figure*}

\textbf{Baselines.} We included the following state-of-the-art counterfactual estimators developed for the multiple treatment setting in the presented experimental evaluation: A composite model consisting of a separate ridge regression model for each possible treatment set $T$ \cite{kallus2017recursive}, a composite ridge regression model \cite{kallus2017recursive} with the input data preprocessed using a Deconfounder \cite{wang2019blessings}, composite models consisting of a separate nearest neighbour predictor (kNN) and Gaussian Process (GP) regressors for each possible treatment set, Treatment Agnostic Regression Network (TARNET) \cite{johansson2016learning} adapted to the multiple treatment setting \cite{schwab2018pm} modelling each possible treatment interaction as an additional, separate treatment path, and Generative Adversarial Nets for inference of Individualised Treatment Effects (GANITE) \cite{yoon2018ganite} with an outcome prediction node $y_T$ for each possible treatment interaction set and loss weights $\alpha = \beta = 1$. In addition to unregularized NCoRE, we also tested a version of NCoRE (Balanced) that employed randomised batch matching to train on the data in a manner that controls for observed confounding using low-dimensional balancing scores.

\textbf{Hyperparameters.} We took an unbiased, systematic approach to hyperparameter search in which we used a fixed hyperparameter optimisation budget of at most 30 hyperparameter optimisation runs for all compared methods. For each of the runs, we selected hyperparameters from pre-defined ranges with uniform probability (\Cref{tb:hyperparameters}). After hyperparameter optimisation, we selected the model with the highest validation set performance in terms of root mean squared error (RMSE) in predicting the factual, observed outcomes on the validation fold for each model type.

\textbf{Metrics.} We evaluate models by calculating the root mean squared error (RMSE) 
\begin{align*}
\text{RMSE}(y, \hat{y}) = \frac{1}{2^{|T_\text{complete}|}-1}\Sigma^{j\in [1 \isep 2^{|T_\text{complete}|}-1]} (y_j - \hat{y}_j)^2
\end{align*}
between the true counterfactual outcomes $y$ and the estimated counterfactual outcomes $\hat{y}$ for all $2^{|T|}-1$ possible combinations of treatments in the set of all available treatments $T_\text{complete}$ across all patients in the test fold. Note that a set of treatments $T$ has $2^{|T|}-1$ possible combinations and that RMSE can only be evaluated in synthetic or semi-synthetic settings where the true underlying outcome generating process is known. We additionally used bootstrap resampling with 100 bootstrap resamples to compute uncertainty bounds for each presented result, and used two-sided Mann-Whitney-Wilcoxon tests to test whether key results met the pre-specified significance level of $\alpha = 0.05$.

\section{Results and Discussion}
\label{sec:results} %

\begin{figure*}[htp!]
    \vspace{-0.45em}
    \includegraphics[width=0.310\linewidth,page=1]{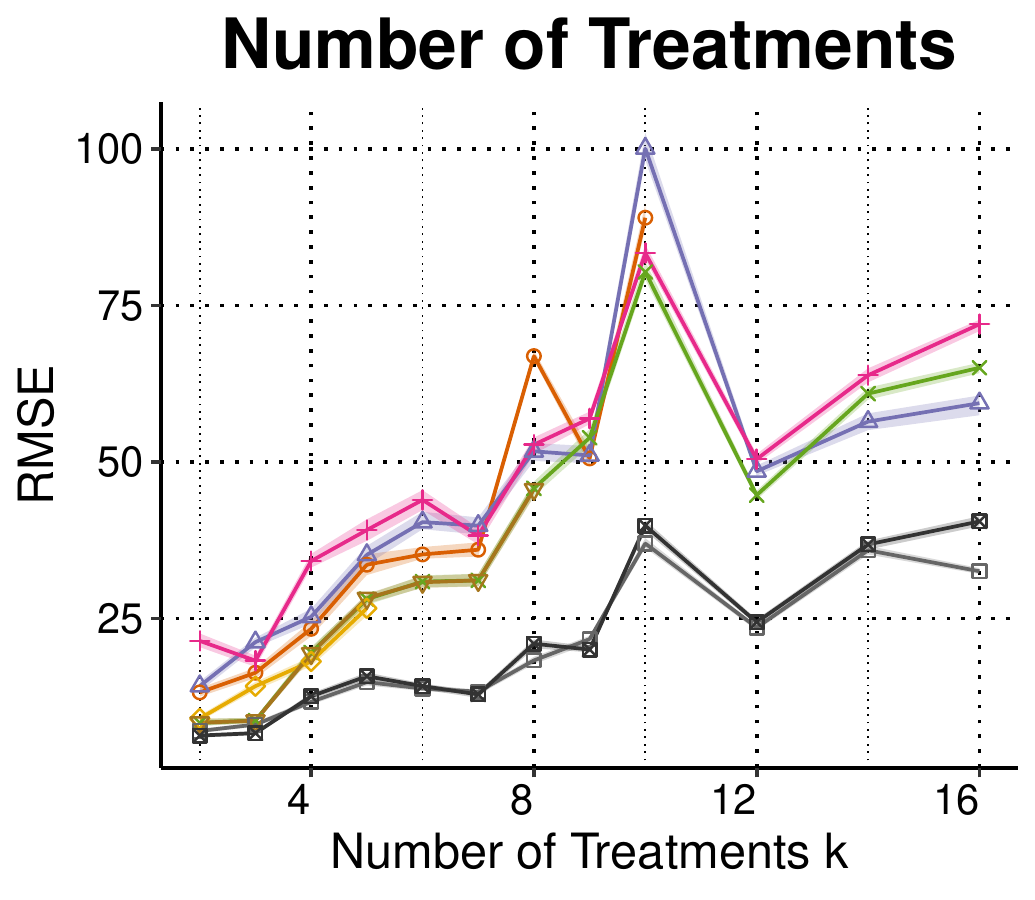}\quad
    \includegraphics[width=0.310\linewidth,page=1]{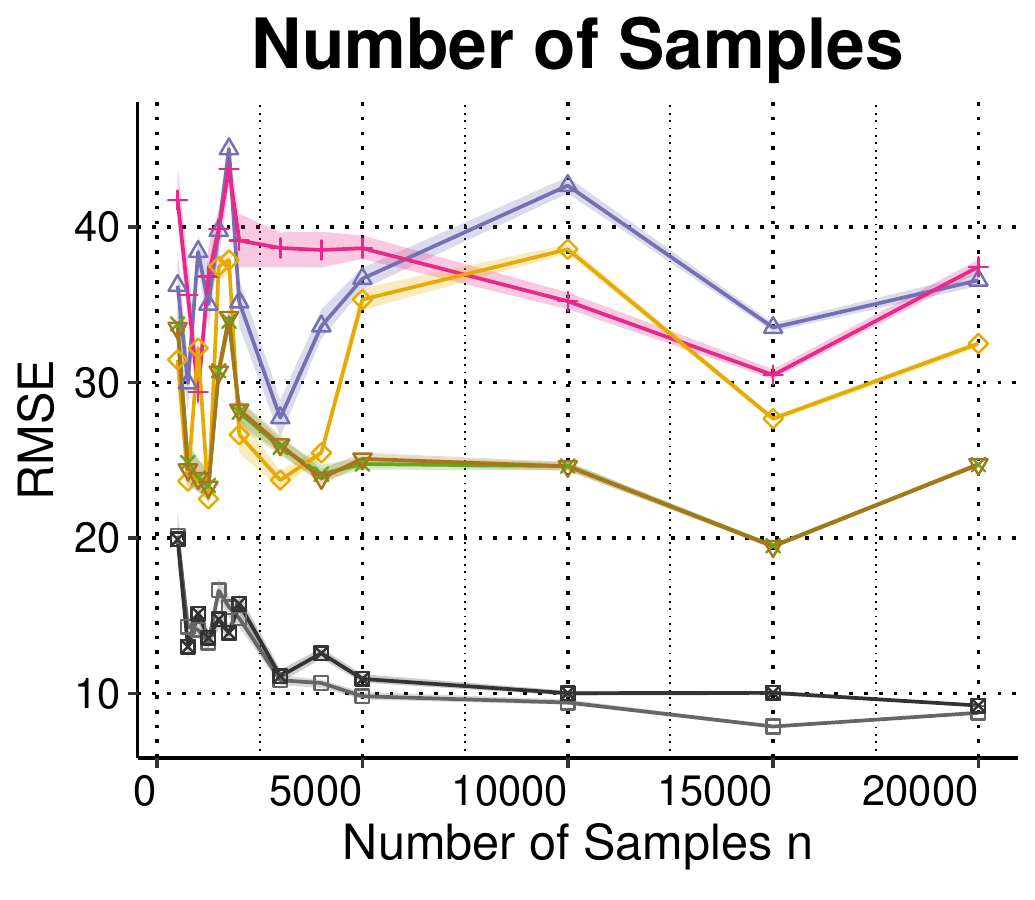}\quad
    \includegraphics[width=0.380\linewidth,page=1]{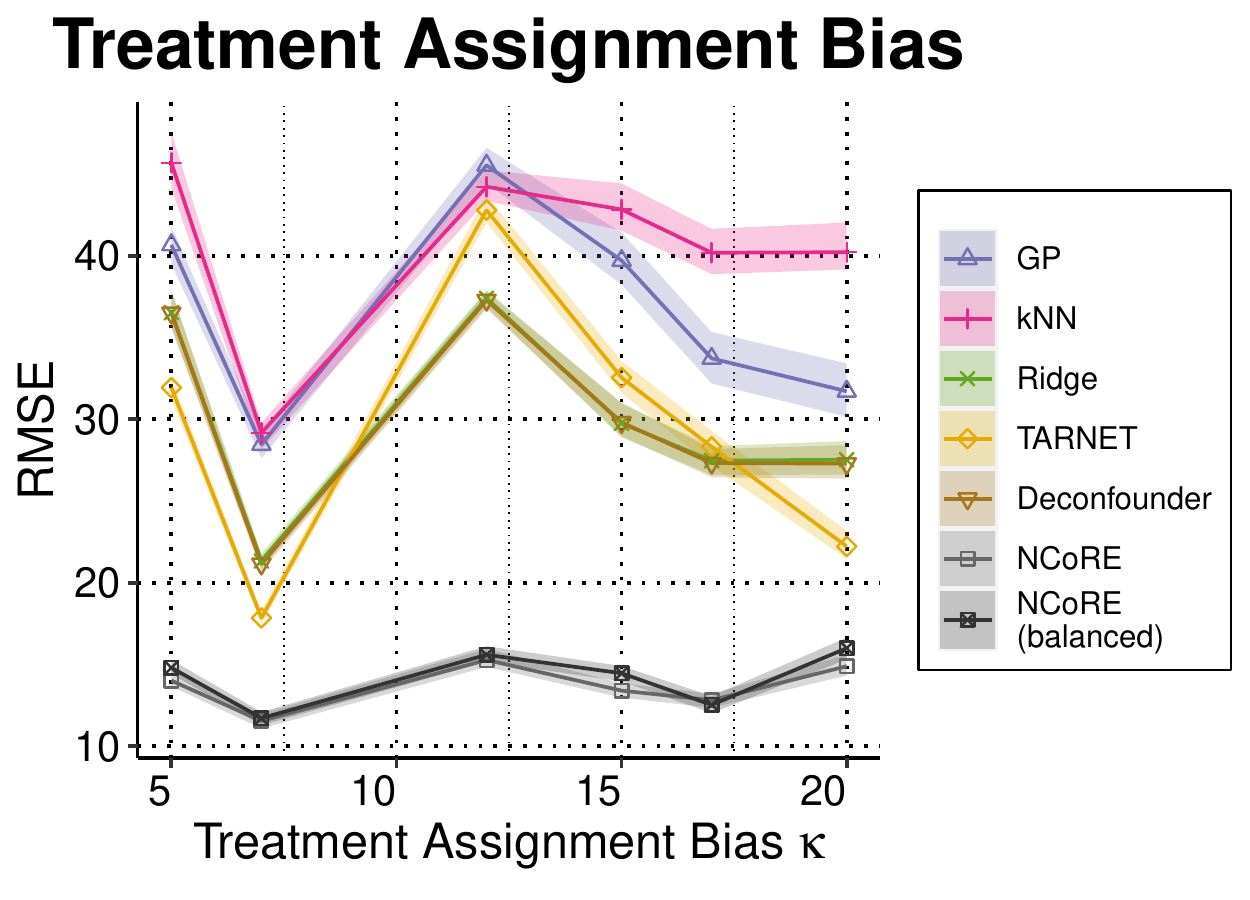}
    \vspace{-2.5em}
    \caption{Performance comparison of several state-of-the-art methods for counterfactual treatment estimation in terms of RMSE (y-axis) for varying levels (x-axis) of treatments $k$ (left), number of samples $n$ (middle) and treatment assignment bias $\kappa$ (right). Overall \themethod{} consistently outperforms all other considered competitive baselines across the entire range of number of treatments, number of samples and treatment assignment bias values.  }
\label{fig:performances_2}
\end{figure*}

\textbf{Counterfactual Estimation Performance.} We compared the performance in terms of RMSE of all the baselines for counterfactual inference on the simulated data, Europe 1, Europe 2 and CRISPR KO (Fig. \ref{fig:performances_1}). Across all the datasets, we find that \themethod{} outperforms all existing state-of-the-art methods in terms of the RMSE. On simulated data there is a significant performance gap to all the baselines where TARNET, RIDGE and Deconfounder perform equally well and at least slightly better than GANITE, GP and kNN baselines. 
On Europe 1, \themethod{} offers the smallest gap in performance improvement compared to all other baselines, which might be explained by the fact that the variance in the results (dotted points on top of the bars) is likewise significantly higher compared to the simulated data and Europe 2. On the real-world CRISPR Three-way Knockout only \themethod{}, kNN and Ridge return results after a reasonable time frame. All methods show performance levels in the same range while balancing seems to have a positive effect on \themethod{}. On the other datasets, the effect of balancing is small and only slightly positive with respect to RMSE on Europe 1. It is thus unclear if randomised batch matching for training on the data whilst controlling for confounding with balancing scores, has a positive effect for \themethod{}. Finally, our results show that  \themethod{} improves upon Deconfounder and other baselines by a large margin across all data sets demonstrating that the interaction subnetworks introduced by \themethod{} are important for learning meaningful representations for counterfactual inference.

\textbf{Performance by Number of Treatments $k$.} To assess the impact of the number of treatments on the performance of \themethod{} we measured the RMSE trained with varying numbers of treatments on the simulated dataset (Fig. \ref{fig:performances_2}). With all other hyperparameters held equal, we found that a larger number of treatments, in general, worsens estimation performance across all baselines. However, \themethod{} has the lowest RMSE regardless of the number of treatments because of its ability to model and account for interactions between treatments. In particular, \themethod{} seems to scale better than the other baselines, since the difference between performance becomes more pronounced as the number of treatments grows. Unlike all the other baselines, by explicitly modelling interactions between treatments using interaction subnetworks, information can be shared that allows \themethod{} to extrapolate more easily to effects of less-frequently occurring treatment combinations as the number of treatments grows.  

\textbf{Performance by Number of Samples $n$.} To determine the efficiency of our model in terms of the influence of the number of samples available for training and evaluation, we compared the performance of \themethod{} against each of the other baselines on the simulated data (Figure \ref{fig:performances_2}). Unlike the other baselines, our method improves in accuracy with an increase in the number of samples. We also find that \themethod{} requires significantly fewer samples for training and evaluation to achieve a lower RMSE. 
We attribute this efficiency to the inclusion of interaction subnetworks in our model that enable knowledge to be shared across common treatments. 

\textbf{Performance by Treatment Assignment Bias $\kappa$.} To assess the robustness of \themethod{} and existing methods to increasing levels of treatment assignment bias in observational data, we compared the performance of \themethod{} to each of the baselines on the simulated test set with varying choices of treatment assignment bias $\kappa \in$  [5, 20] (Fig. \ref{fig:performances_2}). We found that \themethod{} outperformed existing methods across the entire range of evaluated treatment assignment biases.

\textbf{Limitations.} Like other methods that attempt to estimate causal effects from observational data, our method is based on certain assumptions that are untestable in practice. In our work, we assume unconfoundedness which means that our covariate set $X$ contains the most relevant information inferring treatment effects. Without knowledge of the underlying causal process, this is difficult to assess or justify. In addition, we assumed that all treatments have non-zero probability of being observed in the data. In practice, certain combinations may not observable for instance, if the number of available treatments is high, or, in the healthcare setting, if clinical guidelines prevent certain treatments from being applied concurrently.  

\section{Conclusion}
We presented \themethod{}, a new method for learning counterfactual representations to estimate treatment effects of multiple concurrent treatments. Our approach used  a new conditional neural architecture containing treatment interaction subnetworks to model the way in which combinations of treatments may interact when applied in conjunction. Our experiments demonstrated that the model structure is key to learning neural representations for counterfactual inference with multiple treatments from observational data. Overall, \themethod{} significantly outperforms existing state-of-the-art baselines for estimating effects of interventions across several challenging benchmarks, including combination therapy selection in HIV patients and in-vitro causal effect estimation of multi-gene interventions on cancer cell growth based on experimental CRISPR-Cas9 data. Promising future directions include generalising the approach to hidden confounders, extending \themethod{} for modelling sequential treatments, as well as combining observational and interventional data to assess the quality of estimation. Broadly, we see our approach as a means of bridging the long-standing gap between classical methods for reasoning about combination treatment effects and practical applications in medicine, advertising and economics.

\section*{Acknowledgements}
SP is supported by the Swiss National Science Foundation under P2BSP2$\_$184359. PS is an employee and shareholder of GlaxoSmithKline plc.

\bibliography{references}

\begin{thebibliography}{44}
\providecommand{\natexlab}[1]{#1}
\providecommand{\url}[1]{\texttt{#1}}
\expandafter\ifx\csname urlstyle\endcsname\relax
  \providecommand{\doi}[1]{doi: #1}\else
  \providecommand{\doi}{doi: \begingroup \urlstyle{rm}\Url}\fi

\bibitem[Alaa \& van~der Schaar(2017)Alaa and van~der Schaar]{AlaaS17}
Alaa, A.~M. and van~der Schaar, M.
\newblock Bayesian inference of individualized treatment effects using
  multi-task gaussian processes.
\newblock \emph{CoRR}, abs/1704.02801, 2017.

\bibitem[ARCA(2014)]{ARCA}
ARCA.
\newblock Antiretroviral resistance cohort analysis (arca) national database.
\newblock http://www.dbarca.net/, 2014.

\bibitem[Athey \& Imbens(2017)Athey and Imbens]{athey2017state}
Athey, S. and Imbens, G.~W.
\newblock The state of applied econometrics: Causality and policy evaluation.
\newblock \emph{Journal of Economic Perspectives}, 31\penalty0 (2):\penalty0
  3--32, 2017.

\bibitem[Bica et~al.(2020)Bica, Alaa, and Van Der~Schaar]{bica2020time}
Bica, I., Alaa, A., and Van Der~Schaar, M.
\newblock Time series deconfounder: Estimating treatment effects over time in
  the presence of hidden confounders.
\newblock In \emph{International Conference on Machine Learning}, pp.\
  884--895. PMLR, 2020.

\bibitem[Bothwell et~al.(2016)Bothwell, Greene, Podolsky, Jones,
  et~al.]{bothwell2016assessing}
Bothwell, L.~E., Greene, J.~A., Podolsky, S.~H., Jones, D.~S., et~al.
\newblock Assessing the gold standard--lessons from the history of rcts.
\newblock \emph{N Engl J Med}, 374\penalty0 (22):\penalty0 2175--2181, 2016.

\bibitem[Bottou et~al.(2013)Bottou, Peters, Qui{\~n}onero-Candela, Charles,
  Chickering, Portugaly, Ray, Simard, and Snelson]{bottou2013counterfactual}
Bottou, L., Peters, J., Qui{\~n}onero-Candela, J., Charles, D.~X., Chickering,
  D.~M., Portugaly, E., Ray, D., Simard, P., and Snelson, E.
\newblock Counterfactual reasoning and learning systems: The example of
  computational advertising.
\newblock \emph{The Journal of Machine Learning Research}, 14\penalty0
  (1):\penalty0 3207--3260, 2013.

\bibitem[Carpenter(2014)]{carpenter2014reputation}
Carpenter, D.
\newblock \emph{Reputation and power: Organizational image and pharmaceutical
  regulation at the FDA}.
\newblock Princeton University Press, 2014.

\bibitem[Dempster et~al.(2019)Dempster, Rossen, Kazachkova, Pan, Kugener, Root,
  and Tsherniak]{dempster2019extracting}
Dempster, J.~M., Rossen, J., Kazachkova, M., Pan, J., Kugener, G., Root, D.~E.,
  and Tsherniak, A.
\newblock Extracting biological insights from the project achilles genome-scale
  crispr screens in cancer cell lines.
\newblock \emph{BioRxiv}, pp.\  720243, 2019.

\bibitem[DepMap(2020)]{depmap202020q4}
DepMap, B.
\newblock Depmap 20q4 public, Nov 2020.
\newblock URL
  \url{https://figshare.com/articles/dataset/DepMap_20Q4_Public/13237076/1}.

\bibitem[Egami \& Imai(2018)Egami and Imai]{egami2018causal}
Egami, N. and Imai, K.
\newblock Causal interaction in factorial experiments: Application to conjoint
  analysis.
\newblock \emph{Journal of the American Statistical Association}, 2018.

\bibitem[Freedberg et~al.(2001)Freedberg, Losina, Weinstein, Paltiel, Cohen,
  Seage, Craven, Zhang, Kimmel, and Goldie]{freedberg2001cost}
Freedberg, K.~A., Losina, E., Weinstein, M.~C., Paltiel, A.~D., Cohen, C.~J.,
  Seage, G.~R., Craven, D.~E., Zhang, H., Kimmel, A.~D., and Goldie, S.~J.
\newblock The cost effectiveness of combination antiretroviral therapy for hiv
  disease.
\newblock \emph{New England Journal of Medicine}, 344\penalty0 (11):\penalty0
  824--831, 2001.

\bibitem[Green et~al.(2001)Green, Krieger, and Wind]{green}
Green, P.~E., Krieger, A.~M., and Wind, Y.~J.
\newblock Thirty years of conjoint analysis: Reflections and prospects.
\newblock \emph{Interfaces}, 31\penalty0 (3):\penalty0 S56--S73, 2001.
\newblock ISSN 00922102, 1526551X.
\newblock URL \url{http://www.jstor.org/stable/25062702}.

\bibitem[Imai \& Van~Dyk(2004)Imai and Van~Dyk]{imai2004causal}
Imai, K. and Van~Dyk, D.~A.
\newblock Causal inference with general treatment regimes: Generalizing the
  propensity score.
\newblock \emph{Journal of the American Statistical Association}, 99\penalty0
  (467):\penalty0 854--866, 2004.

\bibitem[Imbens(2000)]{imbens2000role}
Imbens, G.~W.
\newblock The role of the propensity score in estimating dose-response
  functions.
\newblock \emph{Biometrika}, 87\penalty0 (3):\penalty0 706--710, 2000.

\bibitem[Johansson et~al.(2016{\natexlab{a}})Johansson, Shalit, and
  Sontag]{johansson2016learning}
Johansson, F., Shalit, U., and Sontag, D.
\newblock Learning representations for counterfactual inference.
\newblock In \emph{International Conference on Machine Learning}, pp.\
  3020--3029, 2016{\natexlab{a}}.

\bibitem[Johansson et~al.(2016{\natexlab{b}})Johansson, Shalit, and
  Sontag]{JohanssonTAR}
Johansson, F.~D., Shalit, U., and Sontag, D.
\newblock Learning representations for counterfactual inference.
\newblock In \emph{Proceedings of the 33rd International Conference on
  International Conference on Machine Learning - Volume 48}, ICML'16, pp.\
  3020--3029. JMLR.org, 2016{\natexlab{b}}.

\bibitem[Kallus(2017)]{kallus2017recursive}
Kallus, N.
\newblock Recursive partitioning for personalization using observational data.
\newblock In \emph{International Conference on Machine Learning}, 2017.

\bibitem[Kallus(2020)]{kallus2020generalized}
Kallus, N.
\newblock Generalized optimal matching methods for causal inference.
\newblock \emph{Journal of Machine Learning Research}, 21\penalty0
  (62):\penalty0 1--54, 2020.

\bibitem[Lechner(2001)]{lechner2001identification}
Lechner, M.
\newblock Identification and estimation of causal effects of multiple
  treatments under the conditional independence assumption.
\newblock In \emph{Econometric evaluation of labour market policies}, pp.\
  43--58. Springer, 2001.

\bibitem[Lopez et~al.(2017)Lopez, Gutman, et~al.]{lopez2017estimation}
Lopez, M.~J., Gutman, R., et~al.
\newblock Estimation of causal effects with multiple treatments: a review and
  new ideas.
\newblock \emph{Statistical Science}, 32\penalty0 (3):\penalty0 432--454, 2017.

\bibitem[Louizos et~al.(2017)Louizos, Shalit, Mooij, Sontag, Zemel, and
  Welling]{Louizos}
Louizos, C., Shalit, U., Mooij, J.~M., Sontag, D., Zemel, R., and Welling, M.
\newblock Causal effect inference with deep latent-variable models.
\newblock In Guyon, I., Luxburg, U.~V., Bengio, S., Wallach, H., Fergus, R.,
  Vishwanathan, S., and Garnett, R. (eds.), \emph{Advances in Neural
  Information Processing Systems 30}, pp.\  6446--6456. Curran Associates,
  Inc., 2017.

\bibitem[Luce \& Tukey(1964)Luce and Tukey]{luce}
Luce, R. and Tukey, J.~W.
\newblock Simultaneous conjoint measurement: A new type of fundamental
  measurement.
\newblock \emph{Journal of Mathematical Psychology}, 1\penalty0 (1):\penalty0 1
  -- 27, 1964.
\newblock ISSN 0022-2496.
\newblock \doi{https://doi.org/10.1016/0022-2496(64)90015-X}.
\newblock URL
  \url{http://www.sciencedirect.com/science/article/pii/002224966490015X}.

\bibitem[McCaffrey et~al.(2004)McCaffrey, Ridgeway, and
  Morral]{mccaffrey2004propensity}
McCaffrey, D.~F., Ridgeway, G., and Morral, A.~R.
\newblock Propensity score estimation with boosted regression for evaluating
  causal effects in observational studies.
\newblock \emph{Psychological methods}, 9\penalty0 (4):\penalty0 403, 2004.

\bibitem[McCaffrey et~al.(2013)McCaffrey, Griffin, Almirall, Slaughter,
  Ramchand, and Burgette]{mccaffrey2013tutorial}
McCaffrey, D.~F., Griffin, B.~A., Almirall, D., Slaughter, M.~E., Ramchand, R.,
  and Burgette, L.~F.
\newblock A tutorial on propensity score estimation for multiple treatments
  using generalized boosted models.
\newblock \emph{Statistics in medicine}, 32\penalty0 (19):\penalty0 3388--3414,
  2013.

\bibitem[Meyers et~al.(2017)Meyers, Bryan, McFarland, Weir, Sizemore, Xu,
  Dharia, Montgomery, Cowley, Pantel, et~al.]{meyers2017computational}
Meyers, R.~M., Bryan, J.~G., McFarland, J.~M., Weir, B.~A., Sizemore, A.~E.,
  Xu, H., Dharia, N.~V., Montgomery, P.~G., Cowley, G.~S., Pantel, S., et~al.
\newblock Computational correction of copy number effect improves specificity
  of crispr--cas9 essentiality screens in cancer cells.
\newblock \emph{Nature genetics}, 49\penalty0 (12):\penalty0 1779--1784, 2017.

\bibitem[Parbhoo et~al.(2017)Parbhoo, Bogojeska, Zazzi, Roth, and
  Doshi-Velez]{parbhoo2017combining}
Parbhoo, S., Bogojeska, J., Zazzi, M., Roth, V., and Doshi-Velez, F.
\newblock Combining kernel and model based learning for hiv therapy selection.
\newblock \emph{AMIA Summits on Translational Science Proceedings},
  2017:\penalty0 239, 2017.

\bibitem[Parbhoo et~al.(2020)Parbhoo, Wieser, Roth, and
  Doshi-Velez]{pmlr-v126-parbhoo20a}
Parbhoo, S., Wieser, M., Roth, V., and Doshi-Velez, F.
\newblock Transfer learning from well-curated to less-resourced populations
  with hiv.
\newblock In Doshi-Velez, F., Fackler, J., Jung, K., Kale, D., Ranganath, R.,
  Wallace, B., and Wiens, J. (eds.), \emph{Proceedings of the 5th Machine
  Learning for Healthcare Conference}, volume 126 of \emph{Proceedings of
  Machine Learning Research}, pp.\  589--609, Virtual, 07--08 Aug 2020. PMLR.
\newblock URL \url{http://proceedings.mlr.press/v126/parbhoo20a.html}.

\bibitem[Robins et~al.(2000)Robins, Hernan, and Brumback]{robins2000marginal}
Robins, J.~M., Hernan, M.~A., and Brumback, B.
\newblock Marginal structural models and causal inference in epidemiology,
  2000.

\bibitem[Rosenbaum \& Rubin(1983)Rosenbaum and Rubin]{propintro}
Rosenbaum, P. and Rubin, D.
\newblock The central role of the propensity score in observational studies for
  causal effects.
\newblock \emph{Biometrika}, 70\penalty0 (1):\penalty0 41--55, 04 1983.
\newblock ISSN 0006-3444.
\newblock \doi{10.1093/biomet/70.1.41}.
\newblock URL \url{https://doi.org/10.1093/biomet/70.1.41}.

\bibitem[Rosenbaum(2002)]{rosenbaum2002overt}
Rosenbaum, P.~R.
\newblock Overt bias in observational studies.
\newblock In \emph{Observational studies}, pp.\  71--104. Springer, 2002.

\bibitem[Rosenbaum \& Rubin(1984)Rosenbaum and Rubin]{rosenbaum1984reducing}
Rosenbaum, P.~R. and Rubin, D.~B.
\newblock Reducing bias in observational studies using subclassification on the
  propensity score.
\newblock \emph{Journal of the American statistical Association}, 79\penalty0
  (387):\penalty0 516--524, 1984.

\bibitem[Schwab et~al.(2018)Schwab, Linhardt, and Karlen]{schwab2018pm}
Schwab, P., Linhardt, L., and Karlen, W.
\newblock Perfect match: A simple method for learning representations for
  counterfactual inference with neural networks.
\newblock \emph{arXiv preprint arXiv:1810.00656}, 2018.

\bibitem[Schwab et~al.(2020)Schwab, Linhardt, Bauer, Buhmann, and
  Karlen]{schwab2020learning}
Schwab, P., Linhardt, L., Bauer, S., Buhmann, J.~M., and Karlen, W.
\newblock Learning counterfactual representations for estimating individual
  dose-response curves.
\newblock In \emph{AAAI}, pp.\  5612--5619, 2020.

\bibitem[Shalit et~al.(2017)Shalit, Johansson, and Sontag]{pmlr-v70-shalit17a}
Shalit, U., Johansson, F.~D., and Sontag, D.
\newblock Estimating individual treatment effect: generalization bounds and
  algorithms.
\newblock In Precup, D. and Teh, Y.~W. (eds.), \emph{Proceedings of the 34th
  International Conference on Machine Learning}, volume~70 of \emph{Proceedings
  of Machine Learning Research}, pp.\  3076--3085, International Convention
  Centre, Sydney, Australia, 06--11 Aug 2017. PMLR.

\bibitem[Spreeuwenberg et~al.(2010)Spreeuwenberg, Bartak, Croon, Hagenaars,
  Busschbach, Andrea, Twisk, and Stijnen]{spreeuwenberg2010multiple}
Spreeuwenberg, M.~D., Bartak, A., Croon, M.~A., Hagenaars, J.~A., Busschbach,
  J.~J., Andrea, H., Twisk, J., and Stijnen, T.
\newblock The multiple propensity score as control for bias in the comparison
  of more than two treatment arms: an introduction from a case study in mental
  health.
\newblock \emph{Medical care}, pp.\  166--174, 2010.

\bibitem[Stuart(2010)]{stuart2010matching}
Stuart, E.~A.
\newblock Matching methods for causal inference: A review and a look forward.
\newblock \emph{Statistical science: a review journal of the Institute of
  Mathematical Statistics}, 25\penalty0 (1):\penalty0 1, 2010.

\bibitem[{Szyma{\'n}ski} \& {Kajdanowicz}(2017){Szyma{\'n}ski} and
  {Kajdanowicz}]{2017scikitmultilearn}
{Szyma{\'n}ski}, P. and {Kajdanowicz}, T.
\newblock {A scikit-based Python environment for performing multi-label
  classification}.
\newblock \emph{ArXiv e-prints}, February 2017.

\bibitem[Wager \& Athey(2017)Wager and Athey]{wager2017estimation}
Wager, S. and Athey, S.
\newblock Estimation and inference of heterogeneous treatment effects using
  random forests.
\newblock \emph{Journal of the American Statistical Association}, 2017.

\bibitem[Wang \& Blei(2019)Wang and Blei]{wang2019blessings}
Wang, Y. and Blei, D.~M.
\newblock The blessings of multiple causes.
\newblock \emph{Journal of the American Statistical Association}, 114\penalty0
  (528):\penalty0 1574--1596, 2019.

\bibitem[Yoon et~al.(2018)Yoon, Jordon, and van~der Schaar]{yoon2018ganite}
Yoon, J., Jordon, J., and van~der Schaar, M.
\newblock {GANITE: Estimation of Individualized Treatment Effects using
  Generative Adversarial Nets}.
\newblock In \emph{{International Conference on Learning Representations}},
  2018.

\bibitem[Zanutto et~al.(2005)Zanutto, Lu, and Hornik]{zanutto2005using}
Zanutto, E., Lu, B., and Hornik, R.
\newblock Using propensity score subclassification for multiple treatment doses
  to evaluate a national antidrug media campaign.
\newblock \emph{Journal of Educational and Behavioral Statistics}, 30\penalty0
  (1):\penalty0 59--73, 2005.

\bibitem[Zazzi et~al.(2012)Zazzi, Incardona, Rosen-Zvi, Prosperi, Lengauer,
  Altmann, Sonnerborg, Lavee, Schülter, and Kaiser]{EuResist}
Zazzi, M., Incardona, F., Rosen-Zvi, M., Prosperi, M., Lengauer, T., Altmann,
  A., Sonnerborg, A., Lavee, T., Schülter, E., and Kaiser, R.
\newblock Predicting response to antiretroviral treatment by machine learning:
  The euresist project.
\newblock \emph{Intervirology}, 55\penalty0 (2):\penalty0 123--127, 1 2012.
\newblock ISSN 0300-5526.

\bibitem[Zhao \& Heffernan(2017)Zhao and Heffernan]{zhao2017estimating}
Zhao, S. and Heffernan, N.
\newblock Estimating individual treatment effect from educational studies with
  residual counterfactual networks.
\newblock \emph{International Educational Data Mining Society}, 2017.

\bibitem[Zhou et~al.(2020)Zhou, Chan, Wan, Yuen, Choi, Li, Tong, Zhong, Sun,
  Bao, Mak, Chow, Khaw, Leung, Zheng, Cheung, Tan, Wong, Chan, and
  Wong]{zhou2020threeway}
Zhou, P., Chan, B.~K., Wan, Y.~K., Yuen, C.~T., Choi, G.~C., Li, X., Tong,
  C.~S., Zhong, S.~S., Sun, J., Bao, Y., Mak, S.~Y., Chow, M.~Z., Khaw, J.~V.,
  Leung, S.~Y., Zheng, Z., Cheung, L.~W., Tan, K., Wong, K.~H., Chan, H.~E.,
  and Wong, A.~S.
\newblock A three-way combinatorial crispr screen for analyzing interactions
  among druggable targets.
\newblock \emph{Cell Reports}, 32\penalty0 (6):\penalty0 108020, 2020.
\newblock ISSN 2211-1247.
\newblock \doi{https://doi.org/10.1016/j.celrep.2020.108020}.

\end{thebibliography}
\bibliographystyle{icml2021}

\end{document}